\documentclass[letterpaper, 10 pt, conference]{ieeeconf} 
\IEEEoverridecommandlockouts                              

\makeatletter
\let\NAT@parse\undefined
\makeatother

\usepackage{amssymb,amsmath}

\usepackage[sort&compress,numbers,sectionbib]{natbib}

\usepackage{cite} 
\usepackage{booktabs}
\usepackage[font=small]{caption}
\usepackage{subcaption}
\usepackage{float}
\usepackage{graphicx}

\usepackage{verbatimbox}
\usepackage{multirow}
\usepackage{balance}
\usepackage{algorithm}
\usepackage{algorithmic}
\usepackage{comment}

\usepackage{dsfont}

\newcommand{\secref}[1]{Sec.~\ref{#1}}
\newcommand{\figref}[1]{Fig.~\ref{#1}}
\newcommand{\tabref}[1]{Table~\ref{#1}}

\usepackage[hyphens]{url}

\usepackage{hyperref}
\hypersetup{bookmarksopen,bookmarksnumbered,
pdfpagemode=UseOutlines,
colorlinks=true,
linkcolor=teal,
anchorcolor=teal,
citecolor=teal,
filecolor=teal,
menucolor=teal,
urlcolor=teal,
breaklinks=true
}
\newtheorem{remark}{Remark}

\usepackage[dvipsnames]{xcolor}
\definecolor{darkblue}{RGB}{0.15,0.15,0.55}
\definecolor{lightgrey}{RGB}{0.75,0.75,0.75}

\definecolor{myred}{RGB}{215,48,39}
\definecolor{myblue}{RGB}{69,117,180}
\definecolor{myorange}{RGB}{252,141,89}
\definecolor{mylightblue}{RGB}{145,191,219}

\definecolor{MYlightblue}{RGB}{217,95,2} 
\definecolor{MYdarkblue}{RGB}{117,112,179} 
\definecolor{MYgreen}{RGB}{27,158,119}

\usepackage{color,soul}
\usepackage{svg}

\newcommand{\xxnote}[3]{}
\ifx\hidenotes\undefined
  \usepackage{color}
  \renewcommand{\xxnote}[3]{\color{#2}{#1: #3}}
\fi

\usepackage[normalem]{ulem}
\useunder{\uline}{\ul}{}

\usepackage{bm}


\title{\LARGE \bf 
MultiPush: Learning to Rearrange with Teams of Car-Like Pushers
}
\author{Jeeho Ahn and Christoforos Mavrogiannis\thanks{Authors are with the Robotics Department, University of Michigan, Ann Arbor, USA. Email: $\{$jeeho, cmavro$\}$@umich.edu}}

\let\oldsubsection\subsection
\renewcommand{\subsection}[1]{%
    \vspace{-3pt}
    \oldsubsection{#1}%
    \vspace{-3pt}
}

\begin{document}

\makeatletter
\let\@oldmaketitle\@maketitle
\renewcommand{\@maketitle}{\@oldmaketitle
   \centering
    \includegraphics[width = \linewidth]{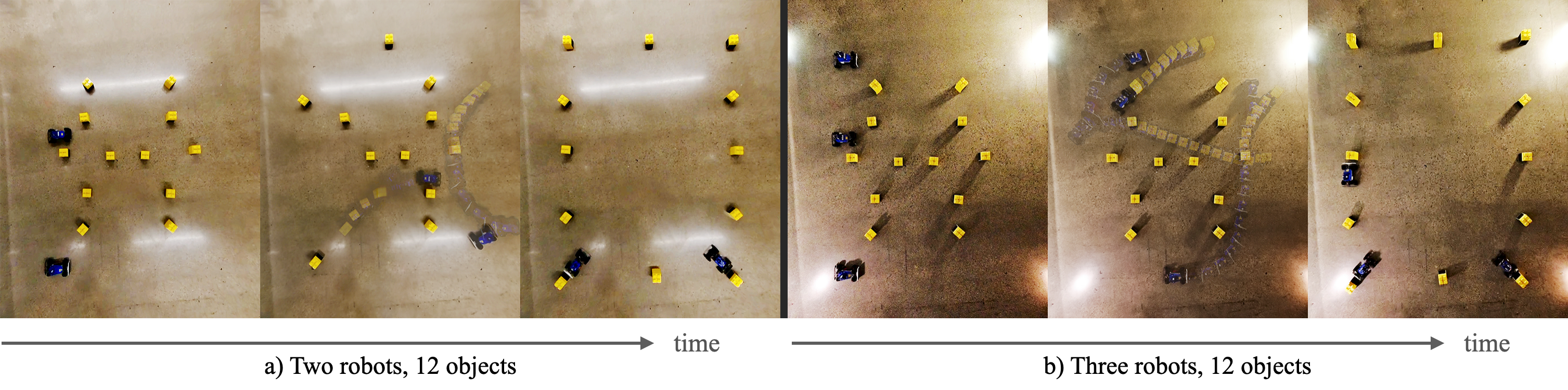}
    \captionof{figure}{We present \emph{MultiPush}, a system for completing multi-object rearrangement tasks in constrained workspaces using a team of car-like robot pushers. At the core of \emph{MultiPush} is a reinforcement learning policy that efficiently assigns robots to pushing tasks. Pictured above are excerpts from hardware experiments using two (a) and three MuSHR robots~\citep{srinivasa2019mushr} (b) on the same 12-object scenario.  Footage from our experiments is available at \url{https://youtu.be/eq-31schnCk}.\label{fig: teaser}}
    }
\makeatother

\maketitle
\thispagestyle{empty}
\pagestyle{empty}

\addtocounter{figure}{-1} 

\begin{abstract}
We focus on the problem of rearranging multiple objects within a constrained workspace via pushing using a team of car-like robots. While the use of multiple robots offers the potential for more efficient execution, the need for conflict resolution and the kinematic constraints arising from physics, robot design, and the workspace boundary make this problem especially challenging. Our key insight is that by exploiting the structure introduced by the car-like kinematics of the domain, we could relax the problem into an ordered assignment of Dubins curves to robots. To this end, we introduce \emph{MultiPush}, a reinforcement-learning based framework that jointly determines an efficient schedule of pushing tasks and their allocation to available robots by leveraging a constraint-aware traversability graph. 
Across extensive simulated trials with up to 14 objects and teams of two to four robots, MultiPush reduces the makespan by up to 16\% compared to the baselines while requiring up to 2.9 times faster planning time. We demonstrate MultiPush on a real-world scenario involving the rearrangement of 12 objects by two and three robots (1/10-scale racecars) in a constrained space.
\end{abstract}


\section{Introduction}\label{sec:introduction}

Teams of autonomous mobile robots are increasingly used for large-scale rearrangement tasks in fulfillment centers, where extensive workspace engineering, including rectilinear rail grids and specialized grippers, makes highly efficient operation possible~\citep{d2012guest}. Such infrastructure, however, is often impractical in inherently unstructured environments, including construction sites, waste-management facilities, and small warehouses. In these settings, dense clutter hinders workspace organization, objects vary widely in shape and size, and robots must operate subject to motion constraints.

Inspired by these real-world challenges, we study the model problem of rearranging multiple objects in a confined workspace using a team of car-like robots that can manipulate objects via pushing rather than grasping. Pushing is a form of nonprehensile manipulation that exploits contact mechanics instead of secure grasps~\citep{lynch1996stable, mason1986mechanics}, enabling robots with simple contact surfaces to move large, heavy, or irregularly shaped objects. However, purposeful physical interaction imposes further motion constraints on robots. Under the quasistatic regime, a car-like robot must steer within a very limited range to ensure stable contact with a manipulated object. Finally, enabling a team of robots to coordinate on simultaneously rearranging multiple objects in a shared, constrained workspace introduces a multirobot coordination challenge: robots must closely follow a carefully designed task schedule, while avoiding conflicts and obeying kinematic constraints.




Our key insight is that the structure imposed by car-like robot constraints during transit to objects and transfer of objects allows us to relax the planning problem from a search over the continuum of multirobot paths into an ordered assignment of Dubins curves~\citep{dubins1957curves,lynch1996stable} to robots. To this end, we introduce \emph{MultiPush}, a planning framework that jointly determines an efficient schedule of pushing tasks and their allocation to available robots via reinforcement learning. MultiPush leverages a traversability graph that captures object connectivity while encoding spatial, kinematic, and physics constraints. We evaluate MultiPush across extensive scenarios involving the rearrangement of up to 14 objects by 4 robots in simulation and show that MultiPush achieves a higher success rate and a lower makespan than baselines while planning the fastest, while scaling to different robot counts and a workspace unseen in training. We also demonstrate MultiPush in the real world with a team of two and three car-like robots rearranging 12 objects.

Our contributions are as follows:
\begin{itemize}
    \item \emph{Multirobot task allocation and scheduling}: We formulate the assignment of the tasks to robots and their execution order as an MDP and train the assignment policy by reinforcement learning on features aggregated over robots and tasks, which can transfer to different number of robots and workspaces.
    \item \emph{Efficient planning}: The learned policy assigns the tasks to robots and their execution order without invoking the motion planner, and the policy is trained to minimize the makespan while avoiding infeasibility.
    \item \emph{Empirical validation}: We validate our framework in simulation on the rearrangement of up to 14 objects with up to four robots, showing the highest success rate and the lowest makespan among the baselines under an equal evaluation budget. The policy generalizes to robot counts and a workspace unseen in training.
\end{itemize}

\begin{figure*}[t]

    \centering
    \includegraphics[width=\linewidth]{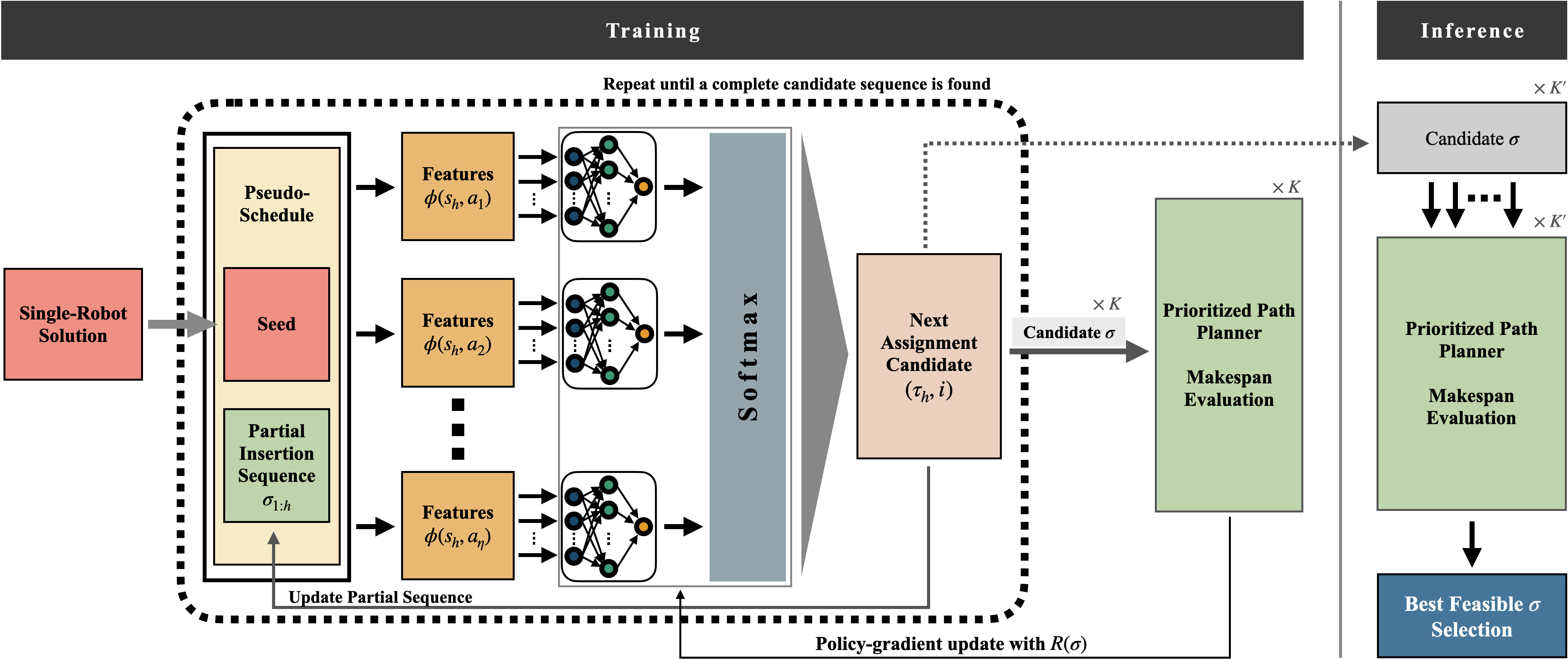}

    \caption{Overview of the MultiPush framework. The single-robot task sequence produced by ReloPush-BOSS~\citep{ahn2026relopushboss} is expanded into \emph{Seed} of resting poses, transit times, task durations, and path traces, which parameterizes a \emph{pseudo-schedule}: a free-space approximation of the multirobot timetable. A learned policy constructs candidate assignment sequences by rolling out against this pseudo-schedule, without invoking the motion planner, and the prioritized multiagent planner verifies them together with the greedy allocation. The best feasible plan is executed.
    $\eta$ denotes the number of possible actions at each step.}

    \label{fig:architecture}
\end{figure*}

\section{Related Work}\label{sec:related-work}



\textbf{Rearranging cluttered workspaces}. While much of the work on rearrangement planning assumes prehensile manipulation via grasping~\citep{stilman2007manipulation,krontiris2015dealing,han2018complexity,ahn2022coordination,ahn2023coordination}, nonprehensile manipulation~\citep{lynch1996stable,mason1986mechanics,goyal1991planar} via pushing empowers robots to handle diverse objects without specialized grippers. Single-robot pushing planners began with clutter removal for object retrieval~\citep{dogar2012planning} and pushing as a pregrasp or rearrangement primitive~\citep{dogar2011framework,king2013pregrasp,king2016rearrangement,haustein2015kinodynamic}. Later work scaled to planar rearrangement of many objects~\citep{huang2019large} and planned object trajectories before projecting them onto feasible robot motions~\citep{ren2025object}. The work of~\citet{ahn2025relopush,ahn2026relopushboss} extends nonprehensile rearrangement to a nonholonomic car-like robots, addressing nonmonotone, densely cluttered instances with a push-traversability graph. Fewer works extend these settings to multiple robots but tend to make strong simplifying assumptions. PuSHR~\citep{talia2023pushr} for example coordinates a team of car-like pushers via multiagent path finding, but is restricted to monotone instances in a discretized workspace. Overall, tackling nonmonotone rearrangement tasks subject to nonholonomic and stability constraints in continuous workspaces is a challenging problem that has been relatively less studied despite its major real-world relevance.

\textbf{Planning for Multiple Agents}. Multiagent Path Finding (MAPF) seeks collision-free paths for multiple agents on shared graphs~\citep{stern2019multi}. Optimal solvers such as Conflict-Based Search (CBS)~\citep{sharon2015conflict} provide completeness guarantees but are computationally expensive in large instances. Prioritized planning offers a scalable alternative where agents plan sequentially, treating higher-priority agents as dynamic obstacles~\citep{erdmann1987multiple,silver2005cooperative,vcap2015prioritized,ma2019searching}. Large Neighborhood Search (LNS) has also been applied to MAPF as an anytime improvement strategy, achieving strong practical performance by iteratively destroying and repairing subsets of agent plans~\citep{li2021anytime,li2022mapf}. Lifelong MAPF assigns incoming tasks online and replans continually, accounting also for kinematic constraints in multiagent pickup and delivery grid-world problems~\citep{MaAAMAS17,ma2019lifelong}. More recently, techniques like reinforcement learning and imitation learning have been employed to enable decentralized coordination across multiple navigating agents: PRIMAL~\citep{sartoretti2019primal} and its lifelong extension PRIMAL2~\citep{damani2021primal} train decentralized MAPF policies by combining imitation and reinforcement learning, and RTAW~\citep{agrawal2023rtaw} allocates warehouse tasks to robots with an attention-based policy trained by reinforcement learning. The path-finding policies move agents on a discrete grid, and the allocation policy scores a candidate by travel delay alone. Less work has studied MAPF problems under kinematic constraints. \citet{wen2022cl} extend CBS to account for car-like kinematics whereas \citet{Davis-RSS-20} introduce arbitrary dynamics into an optimization-based framework. Much of this work is emphasizing navigation settings, lacking mechanisms for handling richer domains like physics-based manipulation. 



\textbf{This work}. We address the centralized coordination of multiple robots for nonmonotone rearrangement tasks under nonprehensile, nonholonomic, and geometric constraints. Prior work has either considered a single pusher, restricted multirobot pushing to monotone instances in discretized spaces, or assigned tasks single-cell to grid-world agents without nonprehensile or nonholonomic constraints. In contrast, our proposed framework, \emph{MultiPush} plans for multiple pushers in a continuous workspace and handles nonmonotone instances. A single-robot plan serves as the seed, fixing how each object is pushed, and a policy trained via reinforcement learning temporally assigns pushing tasks to robots.

\section{Problem Statement}\label{sec:statement}


We consider $n$ identical car-like robot pushers and a set of $m$ rigid,
polygonal objects in a workspace $\mathcal{W}\subset SE(2)$. Each robot
$i\in\mathcal{N}=\{1,\dots,n\}$ has state $p_i\in\mathcal{W}$ and follows
rear-axle simple-car kinematics $\dot{p}_i = f(p_i, u_i)$, where $u_i$ is a
control input (speed and steering angle). Each robot is equipped with a flat
bumper for quasistatic pushing. The state of each object $j\in\mathcal{M}=\{1,\dots,m\}$
is $o_j\in\mathcal{W}$.

Robots start from initial poses $P^s=(p_1^s,\dots,p_n^s)$, and must rearrange all objects from their initial poses
$O^s=(o^s_1,\dots,o^s_m)$ to goal poses $O^g=(o^g_1,\dots,o^g_m)$ by following trajectories $\Xi = (\xi_1,\dots, \xi_n)$, where $\xi_i:[0,1]\to SE(2)$ is the trajectory of robot $i$. We refer to the rearrangement of a single object $j$, moving it from $o^s_j$ to $o^g_j$, as a \emph{task} $\tau_j$, and write $\tau_1,\dots,\tau_m$ for the resulting $m$ tasks, one for each $j\in\mathcal{M}$. We assume an object becomes static once it is rearranged to its goal, and each task can be executed by a single robot acting alone, so that no object requires two robots to move it. A workspace, together with the robots' initial poses and the objects' initial and goal poses, defines a planning \emph{instance}.

An \emph{assignment} pairs a task with the robot that executes it. We refer to the sequence of $m$ assignments, one per task in execution order, as the \emph{assignment sequence} $\sigma$. Our objective is to find an assignment sequence $\sigma$, along with a corresponding tuple of collision-free robot trajectories $\Xi$, that minimizes the \emph{makespan} $C_{\text{max}}$, the completion time of the entire rearrangement. We formalize this problem as:
\begin{equation}
    (\sigma,\,\Xi)^{*} \leftarrow\arg\min_{\sigma,\,\Xi} \;\; C_{\text{max}}[\sigma, \Xi]
    \label{eq:makespan}
\end{equation}

\section{MultiPush: A Framework for Multirobot Nonprehensile Rearrangement Planning}\label{sec:framework}


We present \emph{MultiPush}, a framework for tackling the multi-object rearrangement planning problem with a team of car-like robot pushers described in~\secref{sec:statement}. Our framework comprises a push-traversability graph describing kinematically feasible object transfers and a reinforcement learning policy jointly generating an efficient assignment sequence. An overview of our framework is shown in~\figref{fig:architecture}.

\subsection{Task Representation}

\begin{figure}[t]
    \centering
    \begin{subfigure}[b]{0.49\linewidth}
        \centering
        \includegraphics[width=\linewidth]{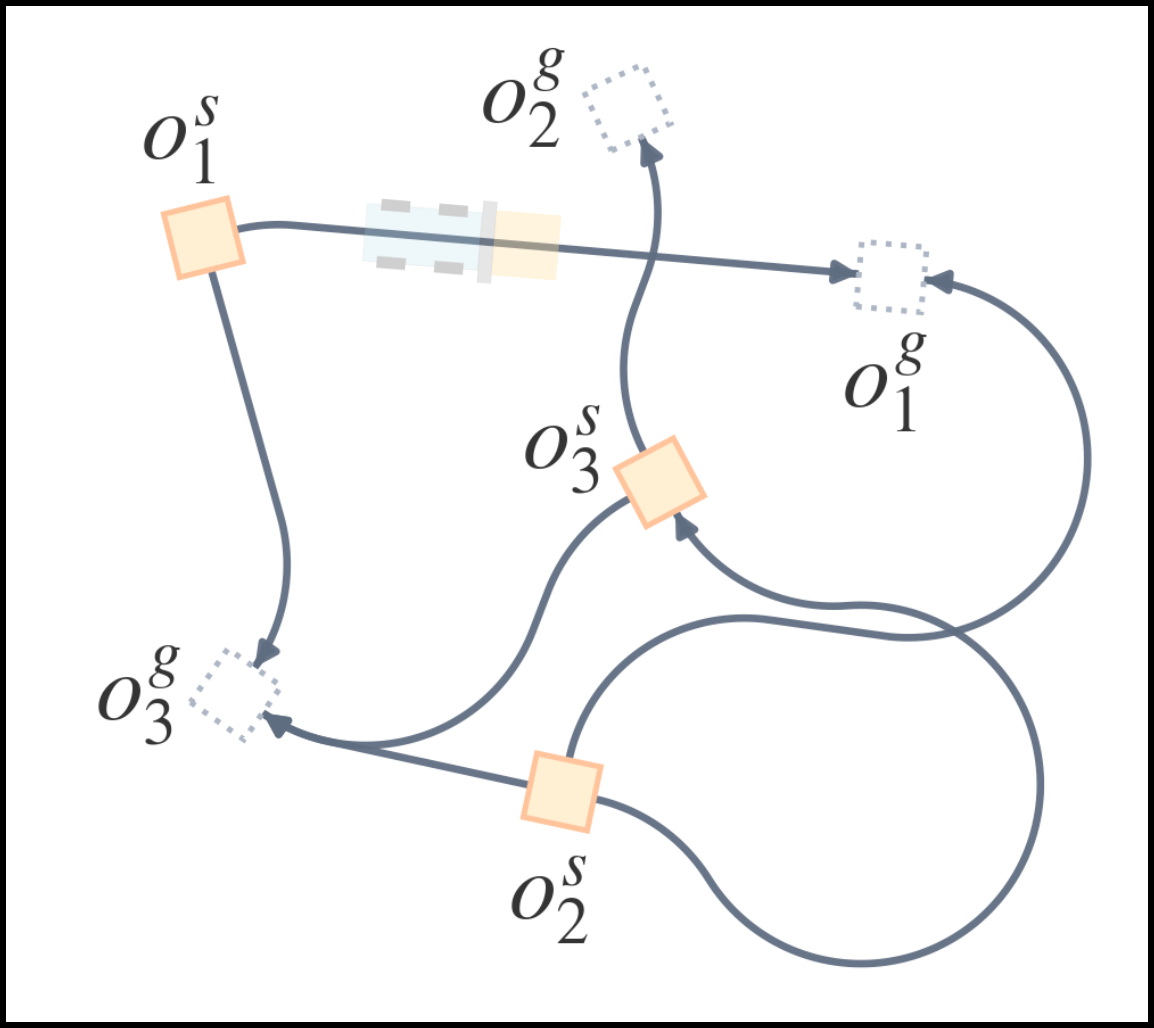}
        \caption{}
        \label{fig:ptgraph-a}
    \end{subfigure}
    \hfill
    \begin{subfigure}[b]{0.49\linewidth}
        \centering
        \includegraphics[width=\linewidth]{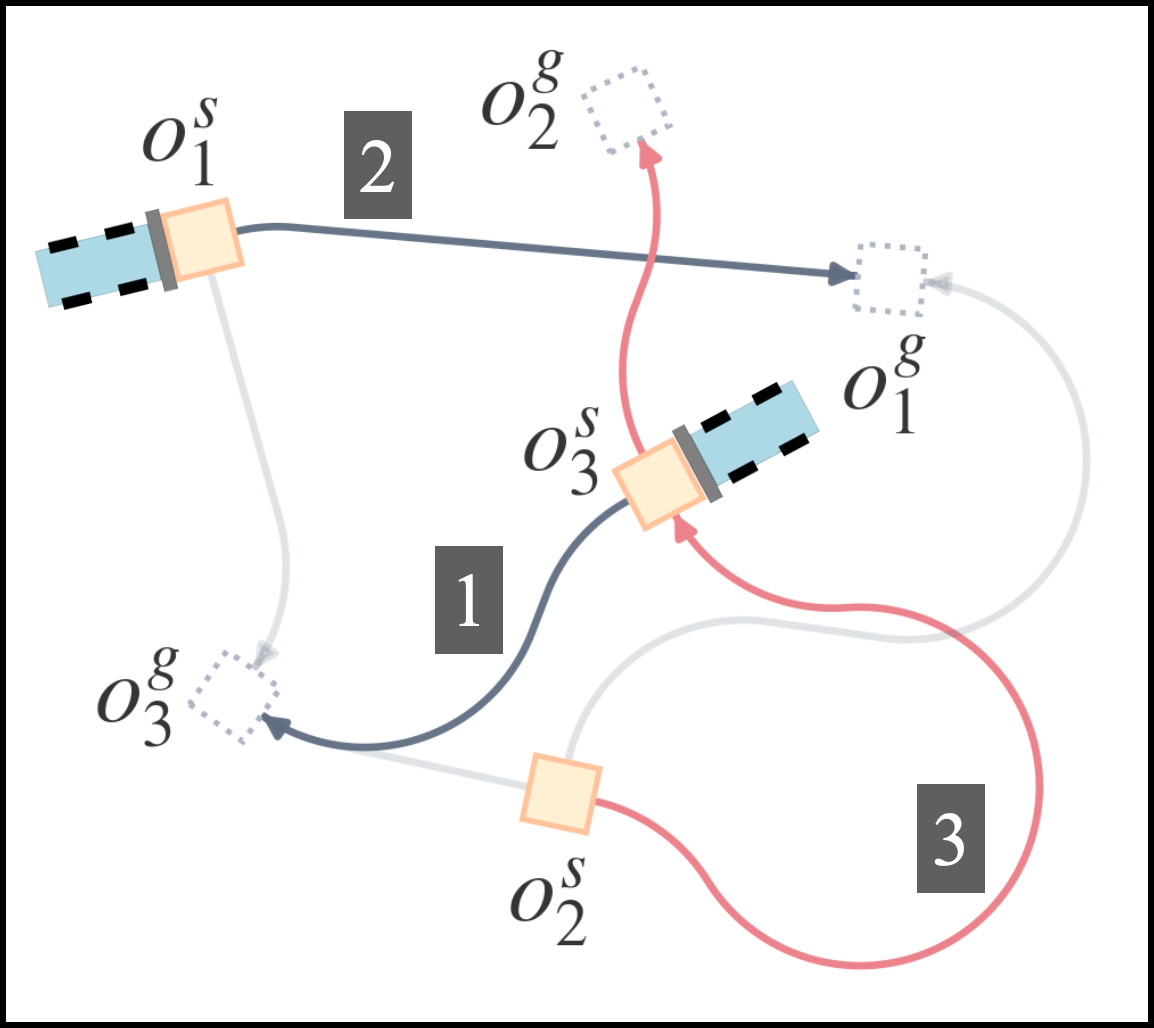}
        \caption{}
        \label{fig:ptgraph-b}
    \end{subfigure}
    \caption{The push-traversability graph. (a) Each vertex represents a start pose or a goal pose at the objects' initial or goal poses, respectively. Each edge represents a feasible transfer path (or a series of partial transfer paths) between the start/goal poses, which are composed of Dubins curves. (b) Feasible rearrangements can be found over graph search at each assignment of tasks to robots. Our goal is to find a feasible assignment sequence with a low makespan.}
    \label{fig:ptgraph}
\end{figure}


Executing a task $\tau_j$ involves a sequence of \emph{transit} and \emph{transfer} paths executed by an assigned robot resulting in the rearrangement of object $j$ from $o^s_j$ to $o^g_j$. The former indicate robot motion without object pushing whereas the latter indicate robot motion while an object is being pushed. For both cases, optimal robot paths can be analytically derived using Dubins curves. To plan a set of robot paths resulting in the rearrangement of all objects, we leverage prior work~\citep{ahn2025relopush,ahn2026relopushboss} on single-robot nonprehensile rearrangement planning. In particular, we design a Push-Traversability graph (PT-graph), comprising \emph{pushing poses} (i.e., robot poses involving robot-object surface contact for objects lying at their initial and goal poses), and edges representing Dubins curves between pushing poses, as shown in~\figref{fig:ptgraph}. By directly searching this graph, we can extract a nominal assignment sequence, $\sigma_0$, comprising a sequence of object transfer paths resulting in the rearrangement of all objects for a single robot pusher~\citep{ahn2026relopushboss,ahn2025relopush}. Leveraging the structure induced by this sequence, we want to determine an efficient task schedule that accommodates $n$ robots.

\subsection{Assignment Sequence as a Markov Decision Process}
\label{sec:mdp}\label{sec:tasks}\label{sec:learning-alloc}


We cast the problem of finding an assignment sequence $\sigma$ as an episodic Markov Decision Process (MDP) with horizon $m$ (the number of tasks to be assigned). Our MDP is a tuple $(\mathcal{S},\mathcal{A},T,R)$, where the state $s_h\in\mathcal{S}$ contains the sequence of assignments already made before step $h$. The action $a_h = (\tau_j, i)$ from the action space $\mathcal{A} = \mathcal{M} \times \mathcal{N}$, taken at step $h$ assigns task $\tau_j$ to robot $i$. The transition function $T$ is deterministic and appends $a_h$ to the state, so that the $m$ actions of a completed episode form the assignment sequence $\sigma$. The reward function $R$ evaluates the makespan resulting from a completed assignment sequence as:
\begin{equation}
  R(\sigma) \;=\;
  \begin{cases}
    -\,C_{\text{max}}(\sigma)
      & \text{if } \sigma \text{ is feasible}, \\[6pt]
    R^{-}(\sigma) & \text{otherwise,}
  \end{cases}
  \label{eq:reward}
\end{equation}
where $C_{\text{max}}(\sigma)$ is the makespan attained by the trajectories computed from collision-free paths generated by a multirobot path planner, normalized per instance by the makespan of the greedy allocation (\secref{sec:experiments}), so that a lower makespan yields a higher reward. The penalty $R^{-}(\sigma)$ follows two constraint-handling rules~\citep{deb2000efficient} where any feasible assignment sequence is preferred to any infeasible one, and infeasible sequences are penalized according to the degree of their violation. We set $R^{-}(\sigma) = R_{\min} - \delta - \lambda\,(1 - h/m)$, where $R_{\min}$ and $\delta$ are the minimum and the standard deviation of the feasible rewards in training, and $h$ is the step at which $\sigma$ first becomes infeasible. The offset $\delta$ places $R^{-}(\sigma)$ strictly below every feasible reward. The term $\lambda\,(1 - h/m)$ ranks the infeasible assignment sequences by how far $\sigma$ progresses before failing. Without the term, every infeasible assignment sequence would receive the same penalty, giving the policy no signal toward candidates that fail later. The weight $\lambda$ bounds the term, keeping the ranking small against the differences among the feasible rewards. Where a single feasible assignment sequence is evaluated with $\sigma$, a constant takes the place of $\delta$, and where there is none, of $R_{\min} - \delta$.

\begin{remark}
    Executing the tasks in an order other than that of $\sigma_0$ is not guaranteed to preserve feasibility: an object at its start or goal placement can block a transfer path planned for another task. An action is thus admissible in state $s_h$ only when its task is not yet assigned and every conflicting task ordered earlier in $\sigma_0$ already is. Admissibility does not guarantee feasibility, which the planner evaluates only once $\sigma$ is complete.

\end{remark}

\begin{remark}
    While $\sigma$ is being filled, the completion time of the task and the next availability of the robot are estimated for each candidate assignment. The estimate accumulates the transit times and task durations of the seed over the tasks already assigned. It plans no paths and therefore accounts for no interaction among the robots. Thus, we refer to this resulting approximate schedule as the \emph{pseudo-schedule} $\hat{\Gamma}_h$, which records which objects have been delivered, where each robot rests, when it next becomes available, and when each scheduled motion occupies its path (\tabref{tab:state}). Evaluating $R(\sigma)$, by contrast, requires planning the trajectories of all robots. This is handled via prioritized planning and explained in detail in~\secref{sec:pathplanning}.

\end{remark}

\begin{table}[t]
\centering
\small
\begin{tabular}{@{}p{0.30\linewidth}p{0.62\linewidth}@{}}
\toprule
Component & Description \\
\midrule
\multicolumn{2}{l}{\textbf{States from the seed (one set per instance)}} \\
Resting poses & robot initial poses and the pose where each task's final transfer path ends \\
Transit times & free-space Reeds--Shepp~\citep{reeds1990optimal} time from each resting pose to each task's pushing pose \\
Task durations & each task's busy time in the seed \\
Path traces & sampled waypoints of each task's paths; used for geometry only \\
Precedence relations & task pairs kept in seed order because a start or goal placement blocks a transfer \\
\midrule
\multicolumn{2}{l}{\textbf{Pseudo-schedule (advanced at each assignment)}} \\
Robot state & resting pose and next availability time of each robot \\
Delivered tasks & objects already at their goals \\
Occupancy windows & when each assigned task uses its path \\
\bottomrule
\end{tabular}
\caption{The geometry, durations, and precedence relations fixed by the seed, and the pseudo-schedule $\hat{\Gamma}_h$ advanced over them (\secref{sec:mdp}). The upper block is computed once per instance from the seed and the robots' initial poses, and the lower block is updated at each assignment. Transit times ignore the other robots, which is the surrogate's central approximation.}
\label{tab:state}
\end{table}

\subsection{Learning an Assignment Policy}


Solving the MDP exactly is impractical, since an instance admits up to $m!\,n^m$ assignment sequences and evaluating one requires planning the trajectories of all robots. We therefore assume a stochastic policy $\pi_\theta$ and train it via reinforcement learning, since no optimal assignment sequences are available to imitate.

For each candidate assignment $a_h$, a feature vector $\phi(s_h, a_h)$ is computed from the seed and the pseudo-schedule $\hat{\Gamma}_h$. The features measure how much the assignment extends the pseudo-schedule, the balance of the robots' availabilities it leaves, and the geometry and ordering that indicate the risk of leaving the remaining tasks infeasible (\tabref{tab:features}). A multilayer perceptron $\psi_\theta$ maps $\phi(s_h, a_h)$ to a score, and a softmax over the scores of the admissible candidates gives the policy $\pi_\theta(a_h \mid s_h) \propto \exp \psi_\theta(\phi(s_h, a_h))$. The same $\psi_\theta$ scores every candidate, so the candidate set may change size at every step and across instances. Because the features are aggregates over robots and tasks or properties of the candidate's own robot, with no entry per robot, the policy applies unchanged to any number of robots.

\begin{table}[t]
\centering
\small
\begin{tabular}{@{}p{0.38\linewidth}p{0.54\linewidth}@{}}
\toprule
Feature & Description \\
\midrule
$\max(0,\ \hat{C}_a-\hat{C}_{\text{max}})$ & extension of the pseudo-schedule \\
$\max(0,\ \hat{C}_{\text{max}}-\hat{C}_a)$ & margin ahead of $\hat{C}_{\text{max}}$ \\
$\mathrm{transit}$ & free-space time to the pushing pose \\
$\hat{C}_i-\min_{i'}\hat{C}_{i'}$ & delay behind the earliest free robot \\
$\max_{i}\hat{C}^{+}_{i}-\min_{i}\hat{C}^{+}_{i}$ & spread of availabilities \\
$\sum_j \mathrm{blockage}_j$ & objects on the corridor \\
$\sum_j \mathrm{corridor}_j \cdot \mathrm{overlap}_j$ & overlap with assigned paths \\
$\sum_i \mathrm{blockage}_i \cdot \mathrm{overlap}_i$ & idle robots on the corridor \\
$\max(0,\ 1-\mathrm{clearance}/r)$ & proximity to the boundary \\
$\#\,\mathrm{blocked\ transits}$ & objects obstructing the transit \\
\bottomrule
\end{tabular}
\caption{Features of $\phi(s_h,a_h)$, evaluated for every candidate assignment from the seed and $\hat{\Gamma}_h$, without path planning. The availability $\hat{C}_i$ is the time robot $i$ completes its last assigned task in $\hat{\Gamma}_h$, and $\hat{C}_{\text{max}}=\max_i \hat{C}_i$ before the assignment. The prediction $\hat{C}_a$ adds the candidate's transit time and task duration from the seed to its robot's availability, and $\hat{C}^{+}_{i}$ are the availabilities after the assignment. Clearance is the smallest boundary distance over the candidate's pushing poses, normalized by the turning radius $r$ (one at the boundary, zero at $r$ or more). The last feature counts earlier-ordered tasks in $\sigma_0$ whose undelivered objects obstruct the non-push segments of the candidate task in the seed.}
\label{tab:features}
\end{table}

\textbf{Training}. The policy is trained via the policy gradient method~\citep{williams1992simple} on the terminal reward. For each instance the policy samples a group of $K$ candidates for $\sigma$, each assignment drawn from $\pi_\theta(a_h \mid s_h)$, and the path planner evaluates them. A candidate's advantage is its reward relative to the group mean, following the shared baseline of POMO~\citep{kwon2020pomo}. Because all candidates of a group solve the same instance, their mean reward serves as the shared baseline: subtracting it removes the reward offset the instance imposes on every candidate, and no value network has to be trained alongside the policy.

\textbf{Inference}. At inference time, the policy samples a group of candidates for $\sigma$, and the prioritized planner verifies each of them. The feasible plan with the minimum makespan is selected for execution.

\subsection{Prioritized Multirobot Path Planning}\label{sec:pathplanning}

Once an assignment sequence $\sigma$ has been determined, robot trajectories are planned one at a time in the order of $\sigma$, following the prioritized paradigm in which each plan treats already planned trajectories as dynamic obstacles in space-time~\citep{erdmann1987multiple,vcap2015prioritized}. A robot assigned a task receives the highest priority, while idle robots receive the lowest. Individual plans are computed via Hybrid~$A^*$~\citep{dolgov2008practical}, which respects the car-like kinematics of the robots and routes around obstacles where a path exists. Planned trajectories are recorded in a centralized \textit{timetable} $\Gamma$, which holds the pose of every robot and object at each discrete timestep and extends the reservation table of cooperative pathfinding~\citep{silver2005cooperative} to objects as well as robots. 

\textbf{Conflict resolution}. Prioritized planning is incomplete in general~\citep{vcap2015prioritized}: a robot can be blocked either by a higher-priority robot still moving toward its destination, or by one that has already reached its destination and remains there. Completeness guarantees exist for \emph{well-formed infrastructures}~\citep{vcap2015prioritized}, environments in which every start and goal location can be reached without passing through any other, so that a resting robot never blocks the path of another. Our setting does not meet this condition, since object goals are fixed by the instance and a robot resting where a push ended may block the path of a later task. We therefore resolve the two blocking cases through the following rules:
\begin{enumerate}
    \item \emph{Wait}: If the blocking robot or object is in motion and will clear the path within a bounded time window, the planner delays the start time.
    \item \emph{Safe parking}: If the conflict is caused by an idle robot, the planner first computes a collision-free trajectory to move the idle robot to a nearby safe parking location, then replans the original path. The method searches for a safe parking by expanding motion primitives from the pose of the blocking robot.
\end{enumerate}
Both apply to every conflict within $\Gamma$'s time horizon up to the point of registering the new trajectory, including conflicts that arise after the robot reaches and rests at its goal. If neither resolves the conflict, the task cannot be scheduled for its assigned robot, and the assignment sequence is declared infeasible.

\section{Evaluation}\label{sec:experiments}

We present an empirical evaluation of MultiPush in simulation, assessing its solution quality, planning efficiency, and generalization to robot counts and a workspace unseen in training.  Footage from our experiments can be found at \url{https://youtu.be/eq-31schnCk}.

\subsection{Experiment Design}

\begin{figure*}[t]
    \centering
    \includegraphics[width=0.8\linewidth]{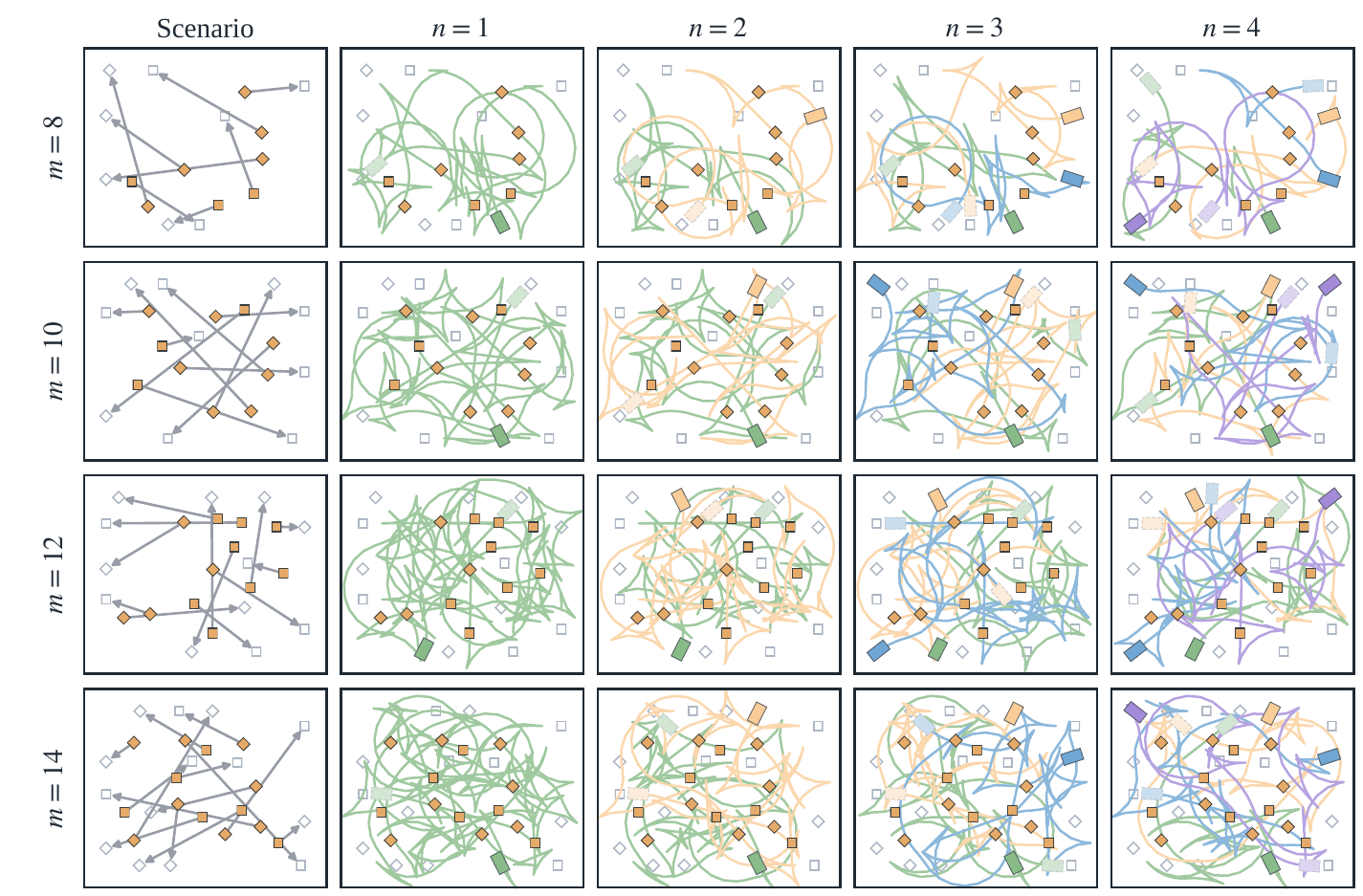}
    \caption{Qualitative results. The first column shows four scenarios drawn from the test set involving respectively $m\in\{8\}$, 10, 12, and 14 objects; objects' start poses are drawn in yellow, and goal poses in white. The remaining columns show the resulting trajectories planned by \emph{MultiPush} for $n = 1$, $2$, $3$, and $4$ robots, colored by robot.}
    \label{fig:scenarios}
\end{figure*}

\textbf{Scenarios}. We evaluate MultiPush on multiple scenarios involving the rearrangement of 8, 10, 12, and 14 objects using a team of 2-4 MuSHR~\citep{srinivasa2019mushr} 1/10-scale car-like robots (see~\figref{fig:scenarios}). We consider cubic objects with a side of $0.15$\,m. We generate unseen instances across two different workspaces: a $4.5\times5.5$\,m$^2$ workspace, in which the policy is trained, and a $5.0\times5.0$\,m$^2$ workspace. Following prior work~\citep{ahn2026relopushboss}, we set the minimum turning radius for stable pushing to $1.43$\,m, which represents a practical tradeoff between push stability and maneuverability. For training, validation, and testing, we generate nominal instances and derive local variants of each by perturbing the poses of the objects and goals. The assignment policy is trained in the $4.5\times5.5$\,m$^2$ workspace only and is used unchanged at inference everywhere else.

\begin{table*}[t]
\centering
\footnotesize
\setlength{\tabcolsep}{3pt}
\begin{tabular}{c @{\hspace{7pt}} l @{\hspace{4pt}} r @{\hspace{10pt}} rrr @{\hspace{14pt}} r @{\hspace{10pt}} rrr}
\toprule
 & & \multicolumn{4}{c}{$4.5\times5.5$\,m$^2$ (trained workspace)} & \multicolumn{4}{c}{$5.0\times5.0$\,m$^2$ (unseen workspace)} \\
\cmidrule(lr){3-6} \cmidrule(lr){7-10}
 & & \multicolumn{2}{c}{Rel. Makespan} & & & \multicolumn{2}{c}{Rel. Makespan} & & \\
\cmidrule(lr){3-4} \cmidrule(lr){7-8}
$n$ & Method & vs.\ GREEDY & vs.\ BOSS & Planning (s) & Success (\%) & vs.\ GREEDY & vs.\ BOSS & Planning (s) & Success (\%) \\
\midrule
\multirow{2}{*}{1}
 & BOSS               & -- & 1.000 & -- & 100.0 & -- & 1.000 & -- & 100.0 \\
\cmidrule(lr){2-10}
 & MultiPush+BOSS     & -- & 0.947 (0.044) & 5.1 (4.7) & 100.0 & -- & 0.945 (0.047) & 5.0 (5.2) & 100.0 \\

\midrule
\multirow{4}{*}{2}
 & GREEDY             & 1.000 & 0.602 (0.064) & -- & 96.0 & 1.000 & 0.593 (0.056) & -- & 93.0 \\
\cmidrule(lr){2-10}
 & RANDOM             & 1.024 (0.133) & 0.613 (0.078) & 11.9 (8.2) & 92.5 & 1.026 (0.130) & 0.606 (0.076) & 11.5 (9.5) & 92.5 \\
 & LNS                & 0.895 (0.077) & 0.536 (0.048) & 21.9 (22.1) & 96.0 & 0.900 (0.070) & 0.532 (0.046) & 19.5 (20.5) & 93.0 \\
 & MultiPush          & \textbf{0.872 (0.085)} & \textbf{0.522 (0.047)} & \textbf{8.7 (8.6)} & \textbf{97.5} & \textbf{0.875 (0.075)} & \textbf{0.517 (0.045)} & \textbf{8.6 (9.4)} & \textbf{96.2} \\
\midrule
\multirow{4}{*}{3}
 & GREEDY             & 1.000 & 0.468 (0.061) & -- & 95.5 & 1.000 & 0.461 (0.062) & -- & 92.2 \\
\cmidrule(lr){2-10}
 & RANDOM             & 1.071 (0.163) & 0.496 (0.071) & 17.7 (12.6) & 93.8 & 1.079 (0.154) & 0.493 (0.074) & 17.5 (15.0) & 94.2 \\
 & LNS                & 0.879 (0.086) & 0.408 (0.046) & 32.2 (33.3) & 95.5 & 0.881 (0.082) & 0.403 (0.050) & 30.6 (31.9) & 92.2 \\
 & MultiPush          & \textbf{0.843 (0.093)} & \textbf{0.391 (0.045)} & \textbf{12.5 (12.3)} & \textbf{98.0} & \textbf{0.847 (0.091)} & \textbf{0.387 (0.049)} & \textbf{13.0 (14.8)} & \textbf{97.0} \\
\midrule
\multirow{4}{*}{4}
 & GREEDY             & 1.000 & 0.407 (0.066) & -- & 95.0 & 1.000 & 0.399 (0.068) & -- & 92.2 \\
\cmidrule(lr){2-10}
 & RANDOM             & 1.099 (0.173) & 0.441 (0.065) & 24.0 (17.2) & 94.5 & 1.109 (0.182) & 0.436 (0.069) & 24.5 (20.7) & 94.5 \\
 & LNS                & 0.866 (0.094) & 0.349 (0.050) & 43.9 (40.5) & 95.0 & 0.869 (0.091) & 0.344 (0.053) & 41.2 (42.3) & 92.2 \\
 & MultiPush          & \textbf{0.840 (0.103)} & \textbf{0.339 (0.050)} & \textbf{17.1 (16.3)} & \textbf{97.5} & \textbf{0.834 (0.102)} & \textbf{0.331 (0.059)} & \textbf{18.1 (21.8)} & \textbf{97.5} \\
\bottomrule
\end{tabular}
\caption{Relative makespan, planning time, and success rate with $n$ robots on the held-out test set for the trained and an unseen workspace. Each workspace column covers 400 test instances, 100 per object count $m \in \{8,10,12,14\}$. The test instances appear in neither the training nor the validation set, and ReloPush-BOSS solves every instance. The relative makespan is measured against GREEDY and against the single-robot makespan of BOSS, both averaged over the instances solved by every method. At $n = 1$, MultiPush+BOSS verifies the BOSS solution in addition to the $K'$ candidates in place of the GREEDY allocation.}
\label{tab:ts_results}
\end{table*}



\textbf{Methods}. We compare the following methods, each evaluating $K' = 20$ candidate assignment sequences per instance:
\begin{itemize}
    \item \emph{GREEDY}. A prioritized allocation strategy that processes the tasks in the order of the seed $\sigma_0$, assigning each to the earliest available robot and breaking ties by the shortest Reeds--Shepp distance to the task's pushing pose. The transit path of the selected robot is planned with the prioritized Hybrid~$A^*$ planner (\secref{sec:pathplanning}), falling back to the next available robot when the plan is infeasible. Each task assignment immediately updates the timetable, so higher-priority assignments constrain subsequent ones. GREEDY is also the reference against which the rewards in training and the relative makespan in evaluation are normalized.

    \item \emph{RANDOM}. A policy that fills each assignment with a uniformly random task and robot, evaluated with the same prioritized planner as the other methods. RANDOM samples and evaluates $K'$ candidates per instance, at most four in parallel, and returns the best feasible assignment sequence.

    \item \emph{LNS}. A Large Neighborhood Search~\citep{shaw1998using} that takes GREEDY as its initial feasible solution and produces up to $K'$ candidates, each by removing one or two assignments and reinserting them. The first assignment to remove is selected by one of three heuristics: random, waiting time, or the assignment that finishes last. The waiting-time heuristic also removes the same robot's assignment with the highest waiting time, if any. Instead of scoring every insertion with a path planner run as in standard LNS, each removed task is reinserted at a position in its neighborhood or at either end of the order, and its robot is chosen at random. The best of three such reinsertions becomes the candidate. The candidates are produced in batches of four evaluated in parallel at each iteration.
    
    \item \emph{MultiPush}. The proposed method, sampling $K'$ candidates from the policy and evaluating GREEDY in addition, as for LNS. The evaluations run at most four in parallel as in the other methods.
\end{itemize}

\textbf{Metrics}. We report the success rate of each method, the fraction of the test instances it solves; the relative makespan, the makespan divided per instance by that of GREEDY and averaged over the instances solved by every method; and the planning time, averaged over all instances. We also report the fraction of planner evaluations that returned infeasible for each method.

\subsection{Implementation}\label{sec:implementation}

\textbf{Software.} MultiPush is implemented in C++, including the prioritized Hybrid~$A^*$ planner and the Dubins and Reeds--Shepp primitives. The score network $\psi_\theta$ has a single hidden layer of eight units, trained in Python with no learning framework beyond the standard library, and loaded by MultiPush at inference. We will release the implementation code upon publication.

\textbf{Prioritized path planner.} The prioritized Hybrid~$A^*$ planner expands motion primitives that include waiting in place, and checks each expansion against the poses recorded in $\Gamma$ at $0.05$\,s substeps. It prunes duplicate states on a grid of $0.2$\,m and $\pi/6$\,rad with a time step of $2.5$\,s, refined near object contact to $0.08$\,m and $1.0$\,s. A search ends when it exhausts its options or reaches its iteration limit, 250 iterations for the main search and 750 for a refinement. The safe parking search expands the same primitives from the pose of the blocking robot, under a limit of 25 iterations.

\textbf{Training}. The policy is trained across instances involving $n=3$ agents only, in two stages. In the first stage, behavior cloning attempts to imitate the GREEDY assignment at each step by minimizing the cross-entropy. This stage stops when the validation cross-entropy no longer improves, after 13 epochs. In the second stage, epochs of REINFORCE~\citep{williams1992simple} follow, with rollout groups of $K = 16$ candidates and a group-mean baseline, path planner evaluations of the 233,640 unique $\sigma$ candidates in total, with $\lambda = 0.2$ in $R^{-}(\sigma)$ (see Eq.\eqref{eq:reward}). For each object count $m\in\{8,10,12,14\}$, we randomly generate 100 nominal training instances, all of which ReloPush-BOSS solves. From each nominal instance, we derive 25 local variants by perturbing the position and orientation of every object and goal with noise drawn uniformly within $\pm 0.05$\,m and $\pm 0.1$\,rad, respectively, resulting in a total of 26 instances (including the nominal case). Thus, in total, the training set comprises $100\times 26 = 2600$ instances per object count, for a total of 10,400 instances. Out of this set, each epoch draws a unique subsample of 400 instances, comprising 100 instances for each object count. A pass over the training set is thus 26 epochs. Training stops when a pass over the training set improves the relative makespan by less than $0.01$ over the validation set.

\begin{figure}[t]
    \centering
    \includegraphics[width=0.99\linewidth]{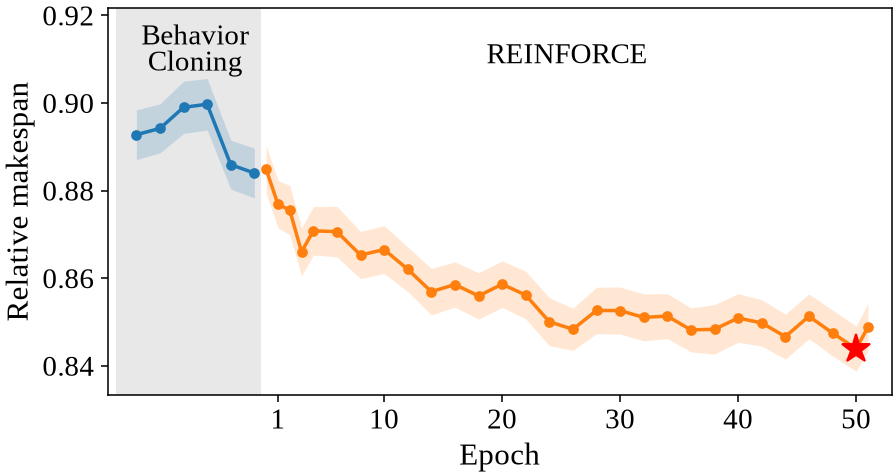}
    \caption{Mean makespan, relative to that of GREEDY, on the validation set across training, warm-started with behavior cloning. Each point averages the relative makespan of the best feasible of the $K'$ sampled candidates over the validation instances the policy solves. The star marks the policy selected for inference. The shaded band is the standard error of the mean. The standard deviation across the solved instances is $0.10$ to $0.12$ at every epoch.}
    \label{fig:training-screen}
\end{figure}

\textbf{Validation}. At each epoch, we sample 480 instances, comprising 40 nominal instances and two variants of each from the trained workspace for a total of 120 instances for each of the 4 object counts (all distinct from the training set). At the end of each epoch, the policy is evaluated at inference: it samples $K'$ candidates, which the prioritized planner evaluates with respect to the GREEDY allocation. The REINFORCE stage stops when a pass over the training set improves the relative makespan by less than $0.01$, after two passes. We select the policy with the lowest relative makespan for inference. \figref{fig:training-screen} shows the policy performance across the behavior-cloning and REINFORCE epochs.

\textbf{Testing}. The policy is tested on 400 instances per workspace, 100 per object count $m$: 20 nominal instances and four variants of each, perturbed as in the training set. The two test sets are generated independently and share no nominal instances. Unlike training, which uses $3$ robots only, each instance is evaluated with $n \in \{2,3,4\}$ robots to characterize generalization to unseen robot counts.


\subsection{Results}\label{sec:results}

\tabref{tab:ts_results} shows the performance of MultiPush against the baselines.

\textbf{Learned policy}. MultiPush achieves a lower relative makespan than LNS in every workspace and robot count, and LNS requires $2.3$ to $2.6$ times the planning time of MultiPush. RANDOM does not improve on GREEDY in any workspace or robot count, so the gains follow from the learned policy rather than from the evaluation budget.

\textbf{Generalization}. MultiPush maintains its advantage over LNS in the unseen workspace and at the two- and four-robot evaluations. The policy therefore generalizes to the robot counts explored and a workspace of different shape unseen in training.

\textbf{Fewer infeasible evaluations}. The planner returns infeasible for a smaller fraction of the evaluations of MultiPush than of LNS and RANDOM (\tabref{tab:infeas_evals}). An infeasible evaluation ends only after the search exhausts its options or reaches its iteration limit, so it consumes the most planning time while yielding no plan. MultiPush thus spends more of the evaluations on plans the robots can execute, which contributes to its planning-time advantage.

\begin{table}
\centering
\footnotesize
\setlength{\tabcolsep}{3.4pt}
\begin{tabular}{l @{\hspace{5pt}} rrr @{\hspace{8pt}} rrr @{\hspace{8pt}} r}
\toprule
\multicolumn{8}{c}{\textbf{Planner evaluations that returned infeasible (\%)}} \\
\midrule
 & \multicolumn{3}{c}{Trained workspace} & \multicolumn{3}{c}{Unseen workspace} & \\
\cmidrule(lr){2-4} \cmidrule(lr){5-7}
Method & $n=2$ & $n=3$ & $n=4$ & $n=2$ & $n=3$ & $n=4$ & All \\
\midrule
RANDOM & 40.6 & 37.7 & 36.3 & 37.7 & 35.3 & 33.6 & 36.9 \\
LNS & 38.7 & 42.0 & 43.4 & 36.4 & 40.6 & 43.2 & 40.7 \\
MultiPush & \textbf{31.2} & \textbf{28.3} & \textbf{26.5} & \textbf{29.1} & \textbf{26.7} & \textbf{24.6} & \textbf{27.7} \\
\bottomrule
\end{tabular}
\caption{Fraction of planner evaluations that returned infeasible on the test set of \tabref{tab:ts_results}.}
\label{tab:infeas_evals}
\end{table}

\section{Hardware Demonstrations}\label{sec:hardware}

We demonstrate MultiPush on a real-world scenario involving the rearrangement of 12 objects using two and three MuSHR 1/10th-scale racecars~\citep{srinivasa2019mushr}, fitted with flat bumpers for pushing (see~\figref{fig: teaser}). Each robot tracks its planned trajectory with a model predictive controller, receiving only its own pose from an overhead motion-capture system. The controller performs path tracking only, with no collision avoidance, since the planned trajectories resolve conflicts among robots and objects (\secref{sec:pathplanning}). The objects are manipulated open loop, with no feedback on their states during execution. Video footage of the demonstrations is provided as a multimedia attachment.

\section{Limitations}\label{sec:limitations}

MultiPush is limited to instances that ReloPush-BOSS~\citep{ahn2026relopushboss} can solve, since the seed fixes the transfer path of every task. While our implementation is adapted from a recent RL framework~\citep{kwon2020pomo}, MultiPush is not tied to it -- other RL frameworks could be swapped out for improved performance. Additionally, the empirical bound of 14 objects also follows from ReloPush-BOSS, whose success rate drops on larger instances. Pushing is modeled as quasistatic, which bounds the robots' steering by the minimum turning radius for stable pushing and leaves slipping and other dynamic effects unmodeled. 
The evaluation covers only identical cubic objects, at most 14 per instance, and two rectangular workspaces of similar size. Adding robots yields diminishing returns, with the fourth robot lowering the makespan less than the third does (see~\tabref{tab:ts_results}), since conflict resolution introduces longer waits as congestion increases. Finally, MultiPush is centralized -- robots are assigned specific trajectories which they then execute independently. In future work, we will extend MultiPush to decentralized execution under limited communication. We will consider objects of different shapes, sizes, and physical properties, and account for contact uncertainty.

\balance
\footnotesize
\bibliographystyle{abbrvnat}
\bibliography{references}



\end{document}